\documentclass{article}

\PassOptionsToPackage{numbers, compress}{natbib}

\usepackage[dblblindworkshop, preprint]{neurips_2025}

\usepackage[utf8]{inputenc} 
\usepackage[T1]{fontenc}    
\usepackage{hyperref}       
\usepackage{url}            
\usepackage{booktabs}       
\usepackage{amsfonts}       
\usepackage{nicefrac}       
\usepackage{microtype}      
\usepackage{xcolor}         
\usepackage{comment}
\usepackage{xspace}
\usepackage{comment}
\usepackage{graphicx}
\usepackage{multirow}
\usepackage{makecell}
\usepackage{amsmath}
\usepackage{subfigure}
\usepackage{newfloat}
\usepackage{listings}
\usepackage{caption}
\usepackage{wrapfig}

\DeclareCaptionStyle{ruled}{labelfont=normalfont,labelsep=colon,strut=off}
\definecolor{bg}{RGB}{250,250,252}
\definecolor{kw}{RGB}{0,86,160}
\definecolor{str}{RGB}{163,21,21}
\definecolor{cm}{RGB}{120,120,120}
\definecolor{idcol}{RGB}{100,0,120}

\newcommand{\qwen}{\textsc{Qwen2.5-VL}\xspace}
\newcommand{\gpt}{\textsc{GPT-4o}\xspace}
\newcommand{\sd}{\textsc{SD-3.5-Medium}\xspace}
\newcommand{\gemma}{\textsc{Gemma-3}\xspace}
\newcommand{\janus}{\textsc{Janus-Pro-7B}\xspace}
\newcommand{\imagen}{\textsc{Imagen-3}\xspace}

\title{The Mind's Eye: A Multi-Faceted Reward Framework for Guiding Visual Metaphor Generation}

\workshoptitle{Generative and Protective AI for Content Creation}

\author{%
  Girish A.\ Koushik\thanks{Correspondence to
    \texttt{g.koushik@surrey.ac.uk}}\\
  {\normalfont NICE Research Group}
  \And
  Fatemeh Nazarieh\\
  {\normalfont NICE Research Group}
  \And
  Katherine Birch\\
  {\normalfont NICE Research Group}
  \And
  \begin{minipage}[t]{\columnwidth}
    \centering
    \begin{tabular}{@{}c@{\qquad}c@{}}
      Shenbin Qian & Diptesh Kanojia\\[0.5ex]
      {\normalfont Centre for Translation Studies} &
      {\normalfont Institute for People-Centred AI}
    \end{tabular}
  \end{minipage}
  \\[3.0ex]
  School of Computer Science \& Electronic Engineering,\\
  University of Surrey, UK\\[1ex]
  \texttt{\{g.koushik,f.nazarieh,k.birch,s.qian,d.kanojia\}@surrey.ac.uk}
}

\begin{document}


\maketitle
\vspace{-0.5cm}

\begin{abstract}
  Visual metaphor generation is a challenging task that aims to generate an image given an input text metaphor. Inherently, it needs \textit{language understanding} to bind a source concept with a target concept, in a way that preserves meaning while ensuring visual coherence. We propose a self-evaluating visual metaphor generation framework that focuses on \textit{metaphor alignment}. Our self-evaluation approach combines existing metrics with our newly proposed metaphor decomposition score and a meaning alignment (MA) metric. Within this setup, we explore two novel approaches: a training-free pipeline that explicitly decomposes prompts into source–target–meaning (S–T–M) mapping for image synthesis, and a complementary training-based pipeline that improves alignment using our proposed self-evaluation reward schema, without any large-scale retraining. On the held-out test set, the training-free approach surpasses strong closed baselines (\gpt, \imagen) on decomposition, CLIP, and MA scores, with the training-based approach close behind. We evaluate our framework output using a user-facing study, and observed that participants preferred \gpt overall, while our training-free pipeline led open-source methods and edged \imagen on abstract metaphors.
  Our analyses show S–T–M prompting helps longer or more abstract metaphors, with closed models excelling on short, concrete cases; we also observe sensitivity to sampler settings. Overall, structured prompting and lightweight RL perform metaphor alignment well under modest compute, and remaining gaps to human preference appear driven by aesthetics and sampling. 
  
  
\end{abstract}
\vspace{-0.5cm}


\begin{figure*}[!t]
    \centering
    \includegraphics[width=0.95\textwidth]{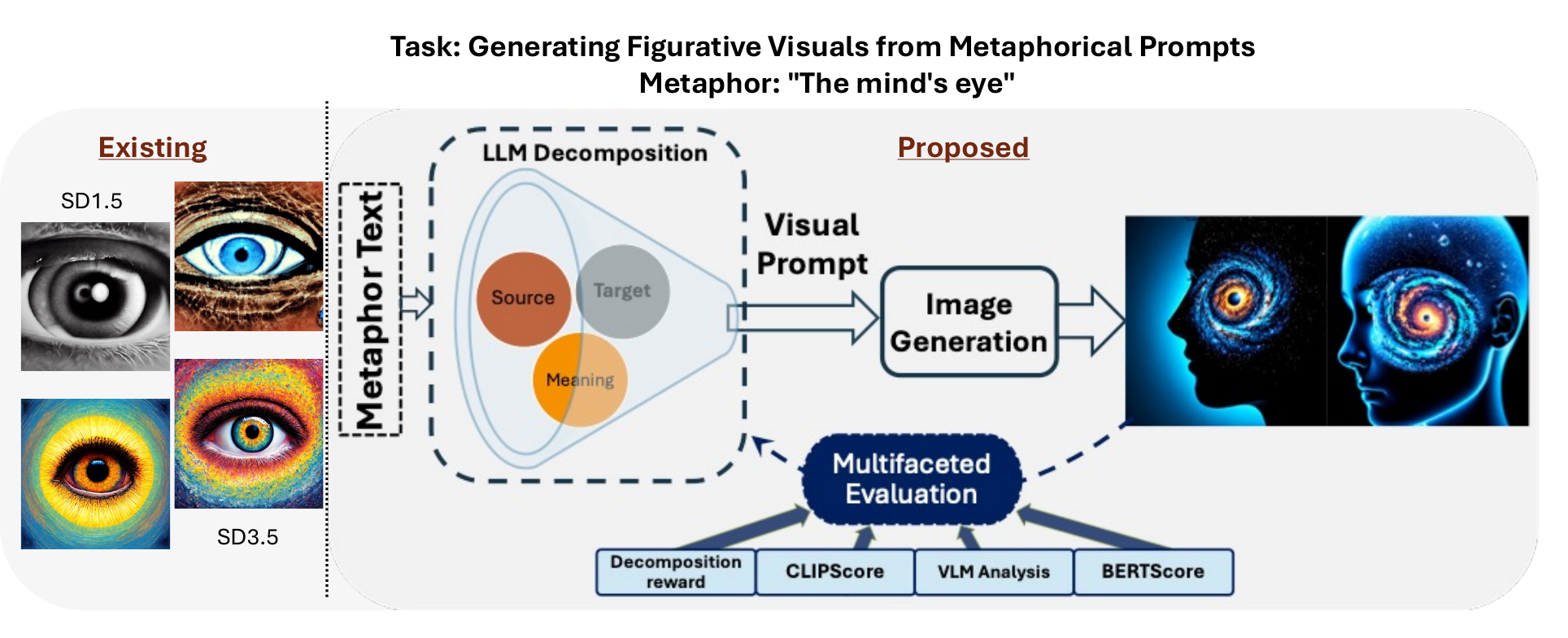}
    \caption{\footnotesize Overview of our framework. Given a metaphorical text input, the pipeline decomposes it into source, target, intended meaning, and visual prompt using a large language model (LLM). A text-to-image model generates an initial image, which is analyzed by a vision-language model (VLM) to assess metaphor alignment. Feedback from the VLM guides iterative refinement in the training-free setting or reinforcement learning fine-tuning (GRPO), resulting in images with improved semantic alignment and metaphorical meaning.}
    \label{fig:graph_abstract}
    \vspace{-0.3cm}
\end{figure*}

\section{Introduction} \label{sec:introduction}
Metaphor is a central cognitive and linguistic phenomenon that shapes how humans conceptualize abstract ideas and communicate complex experiences \citep{Lakoff1980, Johnson1990, Kovecses2010}. By mapping knowledge from a familiar source domain onto a less familiar target domain, metaphors provide a mechanism for reasoning about intangible concepts through concrete experiences. This process not only enriches language but also plays a critical role in persuasion, creativity, and cultural expression \citep{Gibbs2008, Lakoff2008}.

While metaphors are often studied in the context of text, images can offer an even richer medium for figurative expression. Visual metaphors can operate at multiple levels simultaneously, integrating spatial, symbolic, and emotional cues that are difficult to capture through text alone \citep{Forceville1996, Phillips2004}. In multimodal communication such as advertising, political cartoons, and fine art, metaphorical imagery enables rapid, intuitive comprehension by leveraging the human capacity for visual association. Beyond cultural artifacts, effective visual metaphors hold promise for applications in education and scientific communication, where complex concepts benefit from metaphorical illustration.

Recent advances in text-to-image generation have demonstrated the ability to produce photorealistic and stylistically diverse images from natural language prompts \citep{ramesh2022hierarchical, Saharia2022, Balaji2022, Podell2023}. Systems such as DALL·E \citep{ramesh2022hierarchical}, Imagen \citep{Saharia2022}, Stable Diffusion \citep{Rombach2022}, and Midjourney\footnote{\url{https://www.midjourney.com/home}} excel at compositionality and visual fidelity. However, these models remain optimized for literal alignment between prompt and output, often failing to capture figurative intent. For instance, given the metaphor ``the mind’s eye'', most systems simply overlay an image of an eye rather than synthesizing a cohesive visualization that embodies the metaphorical mapping (Figure~\ref{fig:graph_abstract}). Such outputs highlight a fundamental gap between literal surface description and deeper conceptual alignment.

Addressing this gap has proven challenging. Early work has relied on prompt engineering, where large language models (LLMs) expand metaphorical descriptions into more concrete textual prompts to elicit better results from image generators \citep{chakrabarty_i_2023}. While these strategies improve fidelity in some cases, they are typically limited to one-shot rewriting and do not optimize alignment during the generation process. More recent approaches incorporate structured decompositions or fine-tuned retrieval objectives \citep{MetaCLUE, zhang_etal_2024_gome}, but they often lack controllability and rarely integrate iterative feedback from the generated images themselves. As a result, visual metaphors remain an open challenge at the intersection of vision, language, and creativity.

In this paper, we propose a framework for metaphor-aware image generation that explicitly models metaphor structure and integrates evaluation-driven refinement. Drawing from Conceptual Metaphor Theory (CMT) \citep{kramer_conceptual_2025}, our approach decomposes each metaphor into its \emph{source}, \emph{target}, and \emph{intended meaning} (S-T-M), and then guides text-to-image models with feedback from vision language models (VLMs). This design allows us to optimize images not only for visual realism but also for metaphorical alignment. Importantly, we introduce two complementary pipelines: 
\emph{(i) a training-free variant} that leverages the in-context learning abilities of LLMs to iteratively refine prompts without updating model weights, and \emph{(ii) a fine-tuned variant} that applies reinforcement learning (RL) with a multi-faceted reward signal to adapt a smaller LLM for metaphor-aware prompt generation. 

The dual design is significant for two reasons. First, the in-context pipeline requires no retraining or labeled data, making it immediately usable with existing models. Second, the RL fine-tuning variant provides a path toward more specialized and consistent performance without the cost of training from scratch. By fine-tuning only lightweight components, we achieve controllability and improved metaphor alignment while maintaining efficiency.


Our contributions are as follows:
\begin{itemize}
    
    \item \textbf{Metaphor-aware generation pipeline:} We propose a framework that explicitly decomposes each metaphor into its source, target, and intended meaning, and employs iterative feedback to produce images that more faithfully capture figurative intent. To our knowledge, this is the first framework to tightly integrate metaphor theory with text-to-image generation. 
    \item \textbf{Practical dual strategy:} We demonstrate two complementary approaches: (i) a training-free pipeline that requires no model weight updates, enabling immediate deployment under zero retraining cost; and (ii) a reinforcement learning variant that fine-tunes low-rank adapters to achieve stronger and consistent alignment, offering an efficient alternative to training large generative models from scratch.
    \item \textbf{Multi-faceted evaluation and reward design:} We introduce a unified reward function that serves both as an automated evaluation tool and as a tuning signal for refinement and reinforcement learning. Through automatic metrics, analyses, and a human study, we show that our approach displays strong performance in metaphor alignment, especially for longer or more abstract metaphors. 

\end{itemize}



\section{Methodology} \label{sec:methodology}

We propose and evaluate two distinct pipelines for the automated generation of visual metaphors: (1) a training-free approach that relies on the in-context learning capabilities of large language models (LLMs) and (2) a fine-tuning approach that makes LLMs generate descriptions better aligned with the S-T-M of the metaphor. Both pipelines share a common foundation of metaphor decomposition, visual synthesis, and a comprehensive evaluation framework, which we detail below.

\subsection{Core Components} \label{subsec:core-comp}



\paragraph{Metaphor Decomposition.}  The foundational step for both pipelines is the semantic decomposition of the input metaphor (\textit{e.g.}, ``The world is a garden''). The \textit{Source-Target} decomposition is taken from the CMT, and we extend it to include the intended \textit{Meaning} of the metaphor using the source and the target. We define a structured representation consisting of three key elements.
\textbf{Source (S)}: The concrete, familiar domain from which metaphorical meaning is drawn (\textit{e.g.}, ``a garden'').
\textbf{Target (T)}: The abstract or less-understood domain being described (\textit{e.g.}, ``the world'').
\textbf{Meaning (M)}: The intended interpretation or the set of entailments transferred from S to T (\textit{e.g.}, ``The world is a place of potential and beauty that requires cultivation and care''). This S-T-M structure serves as the semantic basis for the generation process. The LLM's primary task is to produce this decomposition and, from it, a \textbf{visual prompt} for the image generation model.


\paragraph{Visual Generation.}  Visual generation is performed by state-of-the-art text-to-image models, including Stable Diffusion (SD) 3.5 Medium \citep{esser2024scaling}, the latest version of a diffusion-based model from Stability-AI and Janus \citep{chen2025janus}, an auto-regressive image generation model from DeepSeek-AI. This component takes the \textbf{visual prompt} generated by the LLM and renders a candidate image. The objective is to create a visual that not only depicts the source but also artistically integrates the abstract qualities of the target and meaning.

\begin{figure*}[!ht]
    \centering
    \includegraphics[width=\linewidth]{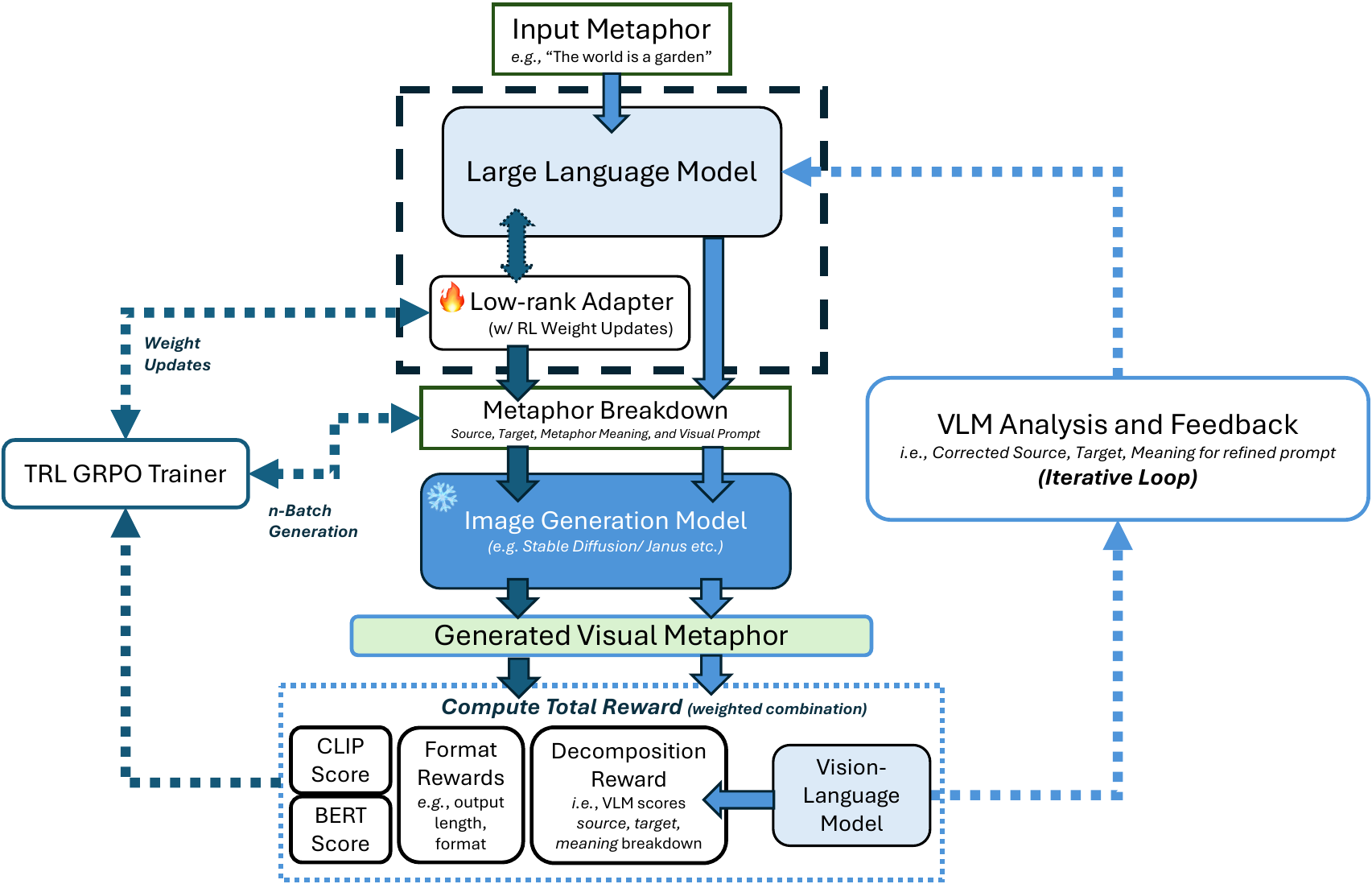}
    \caption{\footnotesize Our proposed Visual Metaphor generation architecture for both training-free and GRPO pipelines. For in-context learning (training-free), we remove the `TRL GRPO Trainer' and provide all the feedback (CLIP, VLM analysis, and BERT scores) through a prompt for refined prompt generation from the LLM. For GRPO, however, the LLM undergoes LoRA fine-tuning with GRPO and the feedback is provided at every training step.}
    \label{fig:vismet-arch}
    \vspace{-0.5cm}
\end{figure*}

\subsection{Multi-Faceted Evaluation Framework} \label{subsec:eval-framework}


A critical novelty of our work is a comprehensive and automated evaluation framework designed to score the quality of a generated visual metaphor. This framework computes several rewards by aggregating multiple quantitative (CLIP, BERT scores) and qualitative (VLM analysis) signals and guides the refinement process in the training-free pipeline and serves as the reward signal for the GRPO fine-tuning process. The framework comprises the following metrics:

\textbf{Decomposition Reward}: A preliminary score, calculated once per metaphor, where a Vision-Language Model (VLM) assesses the semantic quality of the initial S-T-M decomposition itself. This provides a baseline for the coherence of the LLM's analysis.
\textbf{CLIPScore}: We compute the standard CLIPScore~\citep{hessel2021clipscore} to measure the semantic alignment between the generated image and the text of the visual prompt used to create it.
\textbf{VLM-based Analysis}: A VLM (Qwen-VL) is prompted to perform a critical analysis of the generated image, leveraging VLM-as-a-judge~\citep{lee2024prometheus}. It identifies the perceived source ($S^\prime$), target ($T^\prime$), and meaning ($M^\prime$) from the visual content. It then provides scores for the presence and clarity of the source ($vlm_{s\_presence}$), the effectiveness of the symbolic representation of the target ($vlm_{t\_presence}$), and the alignment of the perceived meaning with the original intent ($vlm_{m\_align}$).
\textbf{BERTScore Similarity}: To quantify the semantic fidelity between the original intent and the VLM's perception, we employ BERTScore to calculate the similarity between the original and perceived components: S vs. $S^\prime$ ($bert_{s\_sim}$), T vs. $T^\prime$ ($bert_{t\_sim}$), and M vs. $M^\prime$ ($bert_{m\_sim}$). These scalar rewards provide a holistic measure of metaphor decomposition and the quality of the generated image. Both pipelines are guided by a composite reward function, $\mathcal{R}$, which calculates a total score for a generated image $I$ based on its corresponding visual prompt $P$ and the initial semantic decomposition $D$. The function is a weighted sum of $K$ individual reward metrics:
\vspace{-0.1cm}
\[\mathcal{R}(I, P, D) = \sum_{k=1}^{K} w_k r_k(I, P, D)\]
where $r_k$ is an individual reward function (e.g., CLIPScore, VLM Presence Score, BERTScore).
$w_k$ is the corresponding weight for that metric.


\subsection{Training-Free In-Context Refinement} \label{subsec:in-context}

Our first approach is a zero-shot, training-free pipeline that requires no updates to model weights. It operates as an iterative loop, utilizing the in-context learning abilities of LLMs to refine the visual prompt (shown in Figure \ref{fig:vismet-arch}). The process for a single metaphor is as follows:
\textbf{Initial Decomposition}: The LLM performs the initial S-T-M decomposition and generates a preliminary visual prompt.
\textbf{Iterative Refinement Loop}: For a fixed number of iterations (in our case, 10 iterations): 
\textbf{1.} Generate: An image is synthesized using the current visual prompt. 
\textbf{2.} Evaluate: The image is assessed using the multi-faceted evaluation framework to calculate a Total Reward. 
\textbf{3.} Refine: A new, detailed meta-prompt is constructed for the LLM. This prompt includes the original metaphor, the $S-T-M$ breakdown, the previous visual prompt, and the full suite of evaluation scores and VLM feedback (S', T', M'). The LLM is tasked with analyzing this feedback and generating a revised visual prompt for the next iteration.
\textbf{Selection}: After all iterations are complete, the image from the iteration with the highest \textit{Combined Reward} is selected as the final output.

This method's primary advantage is its flexibility and lack of need for a training dataset or specialized fine-tuning. As a result, it can be directly employed on the test set and the results are discussed in Section~\ref{sec:experiments}. The objective of the training-free pipeline is to find the optimal visual prompt $P^*$ within a fixed number of $N$ iterations, without updating the LLM's parameters. The core idea is to iteratively generate a sequence of prompts and select the one that yields the highest reward. This can be expressed as:
\vspace{-0.2cm}
\[P^* = \underset{P_i \in {P_0, \dots, P_{N-1}}}{\arg\max} \mathcal{R}(\mathcal{G}(P_i), P_i, D)\]
where $P^*$ represents the optimal prompt, $P_i$ is the visual prompt generated at iteration $i$, $P_{i+1}$ is generated by the LLM $\mathcal{L}_{\theta}$ based on feedback from the evaluation of the image $\mathcal{G}(P_i)$, $\mathcal{G}$ is the text-to-image generation model and $D$ is the initial, fixed semantic decomposition of the metaphor.

\subsection{GRPO-based Refinement} \label{subsec:grpo}

Our second approach moves beyond in-context learning to explicitly fine-tune a smaller, more efficient LLM for the task of metaphor decomposition and visual prompt generation. We LoRA-tune \gemma-4B with GRPO~\citep{shao2024deepseekmath} to directly optimize S–T–M with prompt generation for the rewards as described below (Figure~\ref{fig:vismet-arch}).

\textbf{Generation Phase}: For each metaphor in the training set, the model generates multiple candidate responses, each containing a full S-T-M decomposition and a visual prompt.
\textbf{Reward Computation}: Each generated candidate is evaluated using the shared multi-faceted evaluation framework. To optimize this expensive step, the \textit{Metaphor Reward Calculator} class caches results and processes candidates in batches, generating images and performing VLM analysis only once per unique completion.
\textbf{Policy Update}: The GRPO algorithm uses the computed rewards to update the model's LoRA weights. A debiased variant of the GRPO loss is used by setting the loss type to `$dr_{grpo}$' configuration~\citep{liu2025understanding}. This update adjusts the model's policy, increasing the likelihood of generating high-reward outputs and decreasing the likelihood of low-reward ones. By directly fine-tuning on complex, multi-faceted reward signals, this approach aims to instill a more robust and specialized capability for the visual metaphor generation task within a smaller model. The purpose of the GRPO pipeline is to optimize the parameters $\theta$ of the LLM policy $\mathcal{L}_{\theta}$ to maximize the expected reward of its generated outputs. The optimization is driven by minimizing the GRPO loss function, $\mathcal{L}_{\text{GRPO}}$:
\vspace{-0.1cm}
\[\theta^* = \underset{\theta}{\arg\min} \left( -\mathbb{E}{C \sim \mathcal{L}_{\theta}} \left[ \log \sigma \left( \beta \left( \mathcal{R}(C) - \mathcal{R}_{\text{ref}}(C) \right) \right) \right] \right)\]
where $\theta^*$ are the optimal parameters for the LLM, $C$ is a candidate completion (containing a decomposition and prompt) generated from the policy $\mathcal{L}_{\theta}$, $\mathcal{R}(C)$ is the reward for completion $C$, and $\mathcal{R}_{\text{ref}}(C)$ is the reward for the same completion under a reference policy (used for debiasing). $\mathbb{E}$ is `Expectation', which represents the average value of a quantity over a probability distribution, $\beta$ is a hyperparameter controlling the strength of the policy update. $\sigma(\cdot)$ is the sigmoid function.


\paragraph{Experimental Setup.}  All experiments were conducted on dual NVIDIA $A100$ GPUs ($80$GB). We use \gemma~\citep{team2025gemma} as the base LLM, employ \sd~\citep{esser2024scaling} and \janus~\citep{chen2025janus} models for image generation, while \qwen~\citep{bai2025qwen2} served as the vision–language evaluator. Further model specifications and hyperparameters are provided in supplementary material~\ref{sec:exp-setup-app}.

\renewcommand{\arraystretch}{1.5}
\renewcommand{\tabcolsep}{0.1cm}
\begin{wraptable}{r}{0.6\columnwidth}
    \centering
    \caption{\footnotesize Zero-shot generation from input text metaphors. MA: Metaphor Meaning Alignment, S-p: Source-presence, T-p: Target-presence, BERT-S: BERTscore between S and $S^\prime$, BERT-T: BERTscore(T and $T^\prime$), BERT-M: BERTscore(M and $M^\prime$).}
    \resizebox{0.6\columnwidth}{!}{%
    \begin{tabular}{c|cccccccc}
    \toprule
        Model & CLIP Score & MA Score & \makecell{Decomposition \\ Score} & S-p & T-p & BERT-S & BERT-T & BERT-M \\
    \midrule
         \gpt & \textbf{0.2296} & \textbf{0.8180} & 0.8072 & 0.9349 & 0.6856 & 0.8334 & 0.8312 & 0.8823 \\
         \imagen & 0.2224 & 0.7353 & - & - & - & - & - & - \\
    \bottomrule
    \end{tabular}
    \label{tab:zero-shot}
    }%
\end{wraptable}




\paragraph{Zero-Shot Evaluation.}    We begin with zero-shot baselines on the HAIVMet dataset~\citep{chakrabarty_i_2023} using two widely used closed-source, API-accessible models: \gpt~\citep{openai4oimg} and Imagen~\citep{baldridge2024imagen}. \gpt is a natively multimodal model capable of both language and image generation, while Imagen is a diffusion-based image generator without text generation capabilities. For \gpt, we prompt the model to produce not only the image but also textual outputs for the metaphor’s source, target, and intended meaning. This enables evaluation of both visual presence (via VLM analysis of source and target in the generated image), semantic alignment (BERTScore between ground-truth and predicted source/target/meaning), and metaphor decomposition score. For Imagen, which lacks text generation, we restrict evaluation to CLIPScore and meaning alignment computed through VLM analysis.  

\section{Experiments and Results} \label{sec:experiments}


As shown in Table~\ref{tab:zero-shot}, \gpt outperforms Imagen across all shared metrics, achieving a CLIPScore of $0.2296$ and a meaning alignment score of $0.8180$, compared to Imagen’s $0.2224$ and $0.7353$, respectively. These results suggest that multimodal reasoning gives \gpt an advantage for capturing figurative intent, though both models remain limited in producing coherent visual metaphors. This motivates the need for explicit metaphor-aware pipelines introduced in the following sections.

\renewcommand{\arraystretch}{2.0}
\begin{table*}[!ht]
    \centering
    \caption{\footnotesize Results from the training-free and GRPO pipelines. ($^\dagger$): uses an SD guidance scale of 4.5, a resolution of $768\times768$, and 8 inference steps; ($^\ddagger$): uses an SD guidance scale of 1.5, a resolution of $1024\times1024$, and 20 inference steps. MA: Metaphor Meaning Alignment, S-p: Source-presence, T-p: Target-presence, BERT-S: BERTscore between S and $S^\prime$, BERT-T: BERTscore(T and $T^\prime$), BERT-M: BERTscore(M and $M^\prime$).}
    \resizebox{\textwidth}{!}{%
    \begin{tabular}{c|c|c|c|cccccccc}
    \toprule
        Architecture & LLM & \makecell{Image \\ Gen \\ Model} & \makecell{Feedback \\ Model} & \makecell{Decomposition \\ Score} & CLIP Score & MA Score & S-p & T-p & BERT-S & BERT-T & BERT-M \\
    \midrule
        \multirow{4}{*}{GRPO}
         & \gemma-4B & \sd$^\dagger$ & \qwen-32B & 0.7076 & 0.2474 & 0.5266 & 0.9000 & 0.5606 & 0.8161 & 0.7747 & 0.8018 \\      
         & \gemma-4B & \sd$^\ddagger$ & \qwen-32B & 0.6480 & 0.2789 & 0.6396 & 0.9861 & 0.7273 & \textbf{0.8584} & \textbf{0.7987} & \textbf{0.8983} \\     
         & \gemma-4B & Janus-Pro-7B & \qwen-7B & 0.6561 & 0.2916 & 0.6539 & \textbf{0.9874} & 0.7435 & 0.8516 & 0.7762 & 0.8970 \\     
         & \gemma-4B & \sd & \qwen-7B & 0.6480 & 0.2749 & 0.6491 & 0.9829 & 0.7418 & 0.8551 & 0.7936 & 0.8969 \\     
    \midrule
         \multirow{2}{*}{Training-free}
         & \gemma-12B & \sd & \qwen-7B & 0.7212 & 0.2389 & 0.5662 & 0.8910 & 0.5977 & 0.7631 & 0.7562 & 0.8267 \\
         & \gemma-27B & Janus-Pro-7B & \qwen-32B & \textbf{0.8668} & \textbf{0.2960} & \textbf{0.8760} & 0.9058 & \textbf{0.7778} & 0.7676 & 0.7640 & 0.8268 \\
    \bottomrule
    \end{tabular}
    }
    \label{tab:training-results}
\end{table*}

\subsection{Evaluation of Training-Free and GRPO Pipelines} \label{subsec:exp2}
\vspace{-0.3cm}

\begin{figure*}[!ht]
    \centering
    \includegraphics[width=0.99\linewidth]{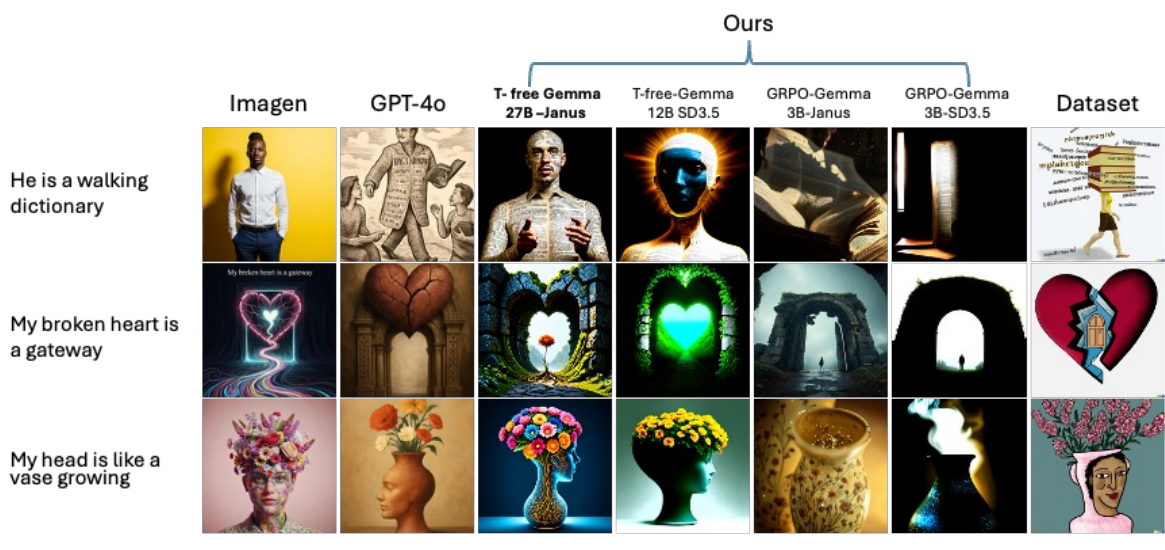}
    \caption{\footnotesize Structure of the user study for evaluating metaphorical image generation. Participants were shown images generated by different methods for the same metaphorical prompt and asked to rate them on metaphor alignment, image quality, and creativity. Ratings were collected and analyzed to compare the performance of zero-shot baselines, training-free pipeline, and GRPO fine-tuning approach. Dataset images are added only for comparison and are not part of the user study.}
    \label{fig:user_study}
    \vspace{-0.1cm}
\end{figure*}



\textbf{Training-free pipeline.} We use \gemma-12B and \gemma-27B as base LLMs, chosen for their strong language understanding and instruction-following abilities. In this setup, VLM analysis is provided as feedback to the LLM, enabling an iterative in-context refinement loop over $10$ iterations. Results (Table~\ref{tab:training-results}) show that the training-free approach substantially improves metaphor alignment: the \gemma-27B + \janus + \qwen-32B configuration achieves the highest CLIP score ($0.2960$), decomposition score ($0.8668$), and MA score ($0.8760$), as well as the best target-presence rate ($0.7778$). These gains underscore the value of larger LLMs and iterative feedback for metaphor-aware generation.  

\textbf{GRPO fine-tuning.} To compare against in-context learning, we apply LoRA-based GRPO fine-tuning on the smaller \gemma-4B, since reinforcement learning is computationally intensive. Despite the reduced capacity, GRPO achieves strong performance: the \gemma-4B + \janus + \qwen-7B configuration obtains a CLIP score of $0.2916$, competitive with training-free results, and the highest source-presence rate ($0.9874$). Performance was further influenced by generation hyperparameters: lowering the guidance scale to $1.5$ and increasing inference steps to $20$ yielded consistently better results compared to standard settings. Both pipelines outperform zero-shot baselines, but with complementary strengths, establishing a foundation for subsequent human evaluation. 

\subsection{User Study} \label{subsec:user-study}

To assess human perception of metaphorical alignment, we conducted a user study with $15$ annotators (all volunteers) on a subset of $50$ metaphors from the test set of $462$ metaphors. An example is provided in Fig.~\ref{fig:user_study}. For each metaphor, participants rated images on \textit{figurative relevance} and \textit{clarity of source–target representation} using a $1$–$5$ Likert scale, and selected the \textit{best overall image} according to semantic alignment and conceptual clarity (rather than style or realism).

\begin{wrapfigure}{r}{0.5\columnwidth}
  \centering
  \subfigure[Best Model]{
    \includegraphics[width=0.49\textwidth]{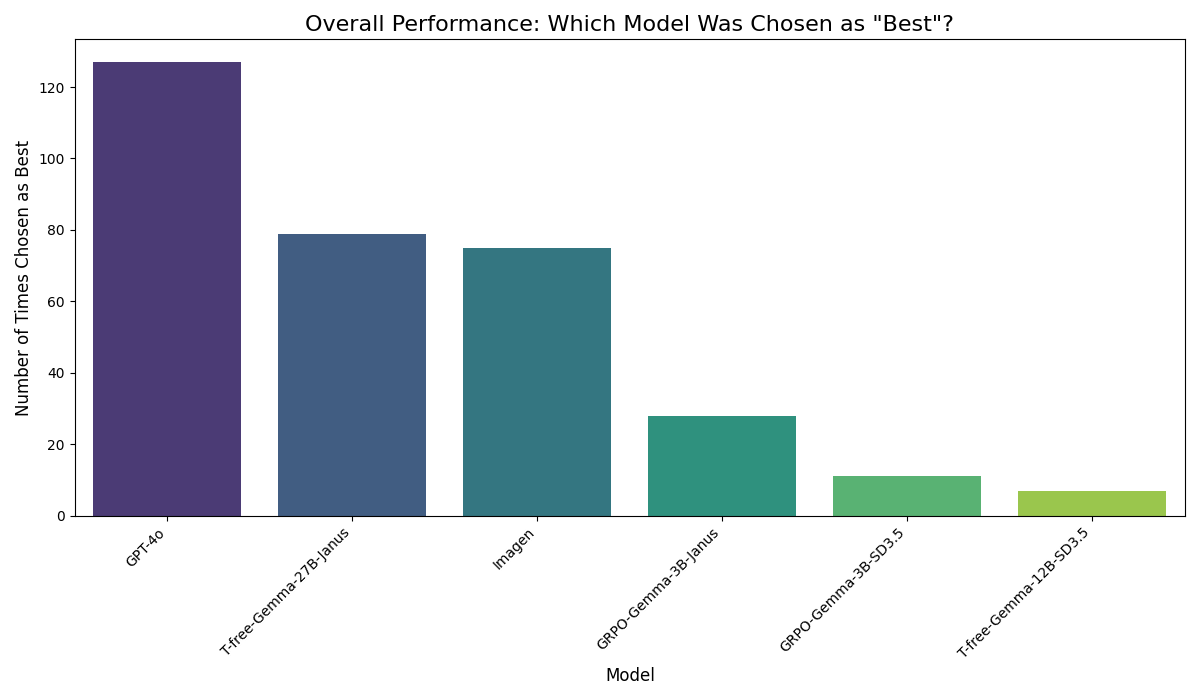}
    \label{fig:best_model}
  }
  \subfigure[Performance by Length]{
    \includegraphics[width=0.49\textwidth]{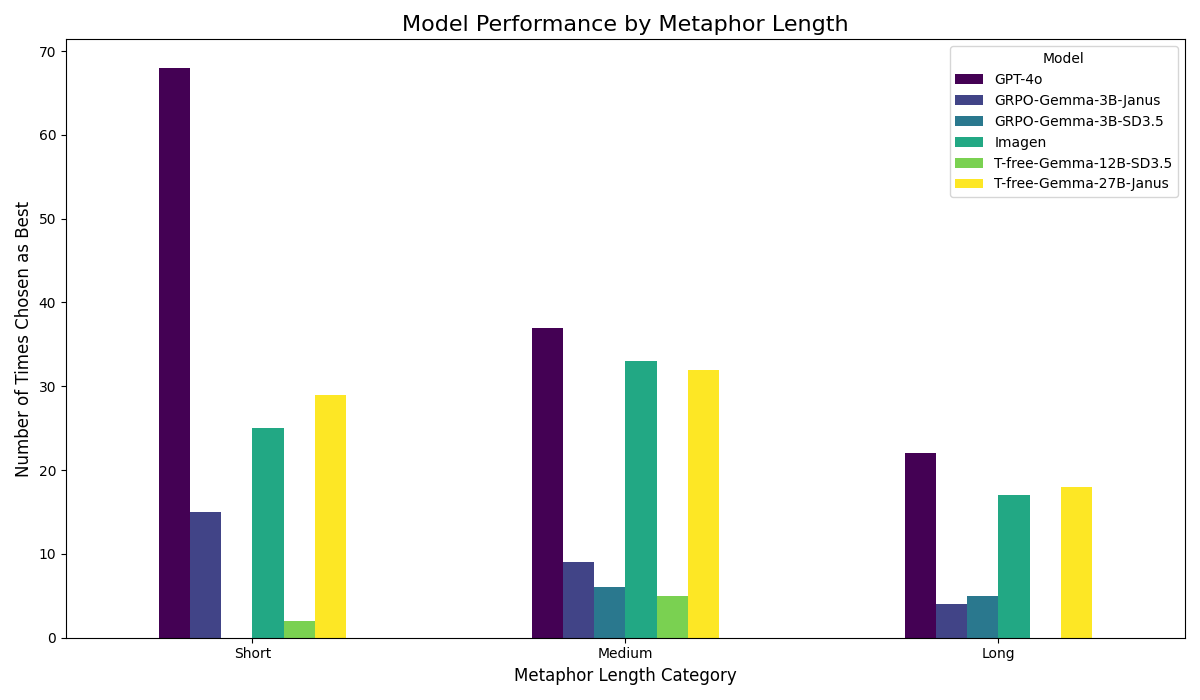}
    \label{fig:len_vs_perf}
    }
  \caption{Best Model Overall and Model Performance w.r.t Metaphor Length}
  \label{fig:best_n_length}
  \vspace{-0.8cm}
\end{wrapfigure}

Results, summarized in Fig~\ref{fig:best_n_length}, reveal a preference hierarchy consistent with automatic metrics. \gpt received the most top votes ($127$), reflecting its strong multimodal backbone. Our training-free pipeline (\gemma-27B + \janus) was the leading open-source system, earning $79$ votes and outperforming \imagen on abstract metaphors. \imagen performed well on concrete categories (\textit{e.g.}, Animals \& Creatures, Plants \& Growth), finishing third ($75$). GRPO-fine-tuned \gemma-4B variants, while trailing in absolute preference ($28$), showed that parameter-efficient fine-tuning can deliver competitive outputs under resource constraints.

\subsection{Discussion} \label{subsec:discussion}
The training-free pipeline achieves the strongest alignment, surpassing both \gpt and \imagen on CLIP, MA, and decomposition scores. GRPO variants, though run on smaller backbones, perform competitively on CLIP and source presence, highlighting their efficiency under limited compute (Tables~\ref{tab:zero-shot},~\ref{tab:training-results}). In contrast, the user study favored \gpt (127 wins) over our training-free pipeline (79) and \imagen (75), with GRPO trailing (28). We attribute this gap to three factors: (i) \gpt’s larger multimodal backbone allows implicit metaphor reasoning, (ii) its highly stylized generations introduce an aesthetics bias despite instructions to judge metaphor faithfulness, and (iii) our study sampled only $50$ of $462$ metaphors, making results sensitive to subset composition. 



Further analysis reveals that \gpt dominates on short, concrete metaphors, while our structured S–T–M refinement narrows the gap on longer or abstract prompts (Figure~\ref{fig:s_and_t}). Category-level results also show complementary strengths: \imagen excels on concrete, visually grounded sources, whereas our system performs best on abstract targets where explicit decomposition helps metaphor alignment. Overall, structured prompting and feedback (training-free) and lightweight reinforcement learning (GRPO) both enable strong metaphor alignment under modest compute. Remaining preference gaps appear driven more by aesthetics and sampling than by semantic understanding, suggesting that style-aware rewards and larger-scale human evaluation are key directions for future work.

\paragraph{Limitations.} Our framework relies heavily on automated model-based scoring (CLIPScore, BERTScore, and a VLM-as-judge) to drive both selection and GRPO rewards. These signals can introduce biases, and they do not fully capture human aesthetic preference, which likely explains the gap between automatic gains and user study wins, as noted in our discussion. The human evaluation itself is modest; 15 annotators on 50 of 462 metaphors, so results may be sensitive to the sampled subset and annotator pool. Performance is also sensitive to generation hyperparameters (e.g., guidance scale and inference steps), indicating tuning effects that may not generalize across setups. Although GRPO shows promise, we only fine-tune a smaller backbone due to compute, so our results may understate the potential of policy optimization at larger scales. Finally, the reward computation and VLM analysis add compute overhead, and our empirical scope is centered on HAIVMet, which limits broader metaphor distributions and domains.

\vspace{-0.1cm}

\section{Related Work}  \label{sec:background}



\paragraph{Visual Metaphor Generation}  Early approaches often relied on prompt engineering with large language models (LLMs) to expand figurative prompts into more concrete textual descriptions for image generation \citep{chakrabarty_i_2023, saakyan_understanding_2025}. While these methods improve over direct prompting, they typically produce overly literal images (e.g., representing ``my bedroom is a pigsty'' with both a messy room and a pig), failing to capture the intended metaphorical meaning. To address this, recent work has explored grounding metaphors in structured decompositions. For example, GOME \citep{zhang_etal_2024_gome} incorporates metaphor grounding and attribute–object binding, showing improvements in semantic alignment. Other efforts, such as MetaCLUE \citep{MetaCLUE}, fine-tune vision–language models for metaphor-aware retrieval, while studies in metaphorical video captioning use structured templates to infuse figurative richness into narrative descriptions \citep{rajakumar_kalarani_etal_2024_unveiling}. However, these methods offer limited controllability and do not directly integrate feedback during the generation process. Our work builds on these directions by explicitly decomposing metaphors into source, target, and intended meaning, and by introducing feedback-driven pipelines to refine metaphor alignment during generation. 
Modern text-to-image models such as DALL·E~\citep{ramesh2022hierarchical}, Imagen~\citep{Saharia2022}, and Stable Diffusion~\citep{Rombach2022} produce high-quality and stylistically diverse images. Recent benchmarks and metrics, including Pick-a-Pic \citep{kirstain2023pickapicopendatasetuser} and TIFA \citep{hu2023tifaaccurateinterpretabletexttoimage}, have improved the evaluation of prompt fidelity. Yet, these models remain largely optimized for literal alignment and surface-level realism, with little focus on the deeper conceptual alignment needed for metaphor illustration.


\paragraph{Reinforcement Learning for Creative Tasks}   Reinforcement learning (RL) has emerged as a powerful tool to enhance creativity and controllability in generative systems. Preference-based approaches such as ImageReward~\citep{xu2023imagereward}, DDPO~\citep{black2023training}, and DPO~\citep{fan2023dpok} leverage human feedback to guide image synthesis, leading to better aesthetics and prompt alignment. More recently, GRPO~\citep{shao2024deepseekmath} has demonstrated strong performance in multimodal generation tasks by improving sample efficiency and stability~\citep{xue2025dancegrpo}. While prior RL methods focus on visual quality or user preference, little work has explored reward-guided optimization for figurative alignment. In this research, we apply GRPO to fine-tune LLMs for metaphor-aware prompt generation, complemented by a training-free iterative refinement pipeline. Together, these approaches directly target the gap between literal image generation and metaphorical meaning alignment.

\vspace{-0.1cm}

\section{Conclusion} \label{sec:conclusion}
We presented a metaphor-aware image generation framework that combines structured S–T–M decomposition with a multi-faceted automated reward to drive both training-free, in-context prompt refinement and parameter-efficient GRPO fine-tuning. By integrating decomposition quality, VLM-based assessments, CLIP, and BERT scores, our approach updates prompts or policies without human intervention, achieving strong meaning alignment. Experiments demonstrate that our methods outperform baselines and offer favorable trade-offs between quality and efficiency, with user studies supporting the automated evaluation. Future work will focus on improving the handling of complex metaphors and closing the remaining gap by unifying alignment and aesthetics in the training signal.

\bibliographystyle{abbrvnat}
\bibliography{neurips_2025}


\appendix

\section{Implementation Details} \label{sec:exp-setup-app}
In this section, we describe the specific experimental setup, models, and hyperparameters used to implement and evaluate our proposed pipelines.

\subsection{Hardware and Software Environment}

All experiments were conducted on a system equipped with two NVIDIA $A100$ GPUs ($80GB$) with a training time of $\sim24$ hours (GRPO) and inference time of $\sim2s$ (including training-free) per metaphor per iteration. Our implementation is built on PyTorch and leverages several high-level libraries from the Hugging Face ecosystem, including Transformers for model access, Diffusers for image generation, and the trl\footnote{\url{https://huggingface.co/docs/trl/main/en/index}} library for our GRPO implementation. To optimize the fine-tuning process for the text LLM, we utilized the unsloth\footnote{\url{https://docs.unsloth.ai/}} library for efficient, memory-saving LoRA training. The Vision-Language Model used for feedback was deployed on a local server using vLLM\footnote{\url{https://docs.vllm.ai/en/latest/}} to ensure high-throughput inference.


\subsection{Model Configuration}

For the GRPO fine-tuning pipeline, we use `\gemma-4B-it-bnb-4bit', a 4-bit quantized version of Google's \gemma-4B model~\citep{team2025gemma}. Visual synthesis was performed using the 4-bit quantized version of `stable-diffusion-3.5-medium` model~\citep{esser2024scaling} from tensorart/stable-diffusion-3.5-medium-turbo\footnote{\url{https://huggingface.co/tensorart/stable-diffusion-3.5-medium-turbo}}, which enables high-quality image generation in fewer steps. Standard generation parameters were set to a guidance scale of $4.5$ and $8$ inference steps, producing images at a resolution of $768\times768$ pixels. The core of our automated evaluation framework is a powerful VLM. We employed the `\qwen-32B-Instruct-AWQ'~\citep{bai2025qwen2} model, a 32-billion parameter, quantized vision-language model, to provide critical analysis and scoring for the generated images.

\subsection{GRPO Fine-Tuning Parameters}

The GRPO-based policy optimization was configured with the following hyperparameters. We used LoRA with a rank of $32$ and an alpha of $64$. The learning rate was set to $5e-5$ with a training batch size of $4$. The maximum sequence length for the LLM generations was capped at 512 tokens, and the number of LLM generations is set to $4$. We utilized the debiased GRPO loss function ($dr_{grpo}$) as implemented in the trl library.

The reward signals that guide both in-context refinement and GRPO training are weighted. A weightage of $0.20$ was assigned to both decomposition reward and CLIP score. The remaining metrics were assigned a weight of $0.10$. For the GRPO pipeline, two additional intrinsic rewards for generation quality, format reward, and length reward were each given a weight of $0.10$ to encourage well-formed and concise outputs.

\section{User study}    \label{sec:appendix_us_imgs}

\begin{figure}[!ht]
  \centering
  \subfigure[Performance Compared to Source Concept]{
    \includegraphics[width=0.45\textwidth]{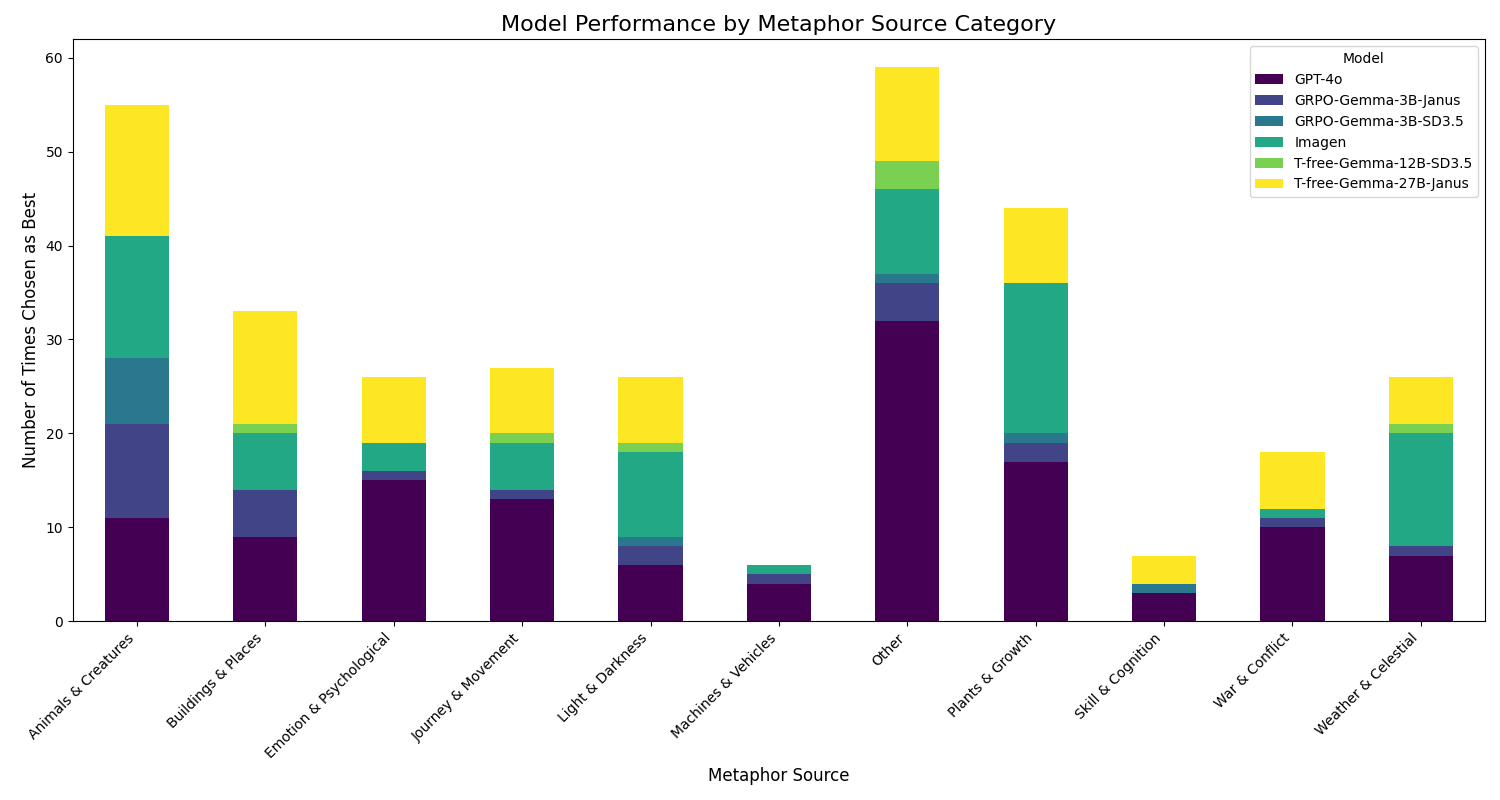}
    \label{fig:source_perf}
  }
  \subfigure[Performance Compared to Target Concept]{
    \includegraphics[width=0.45\textwidth]{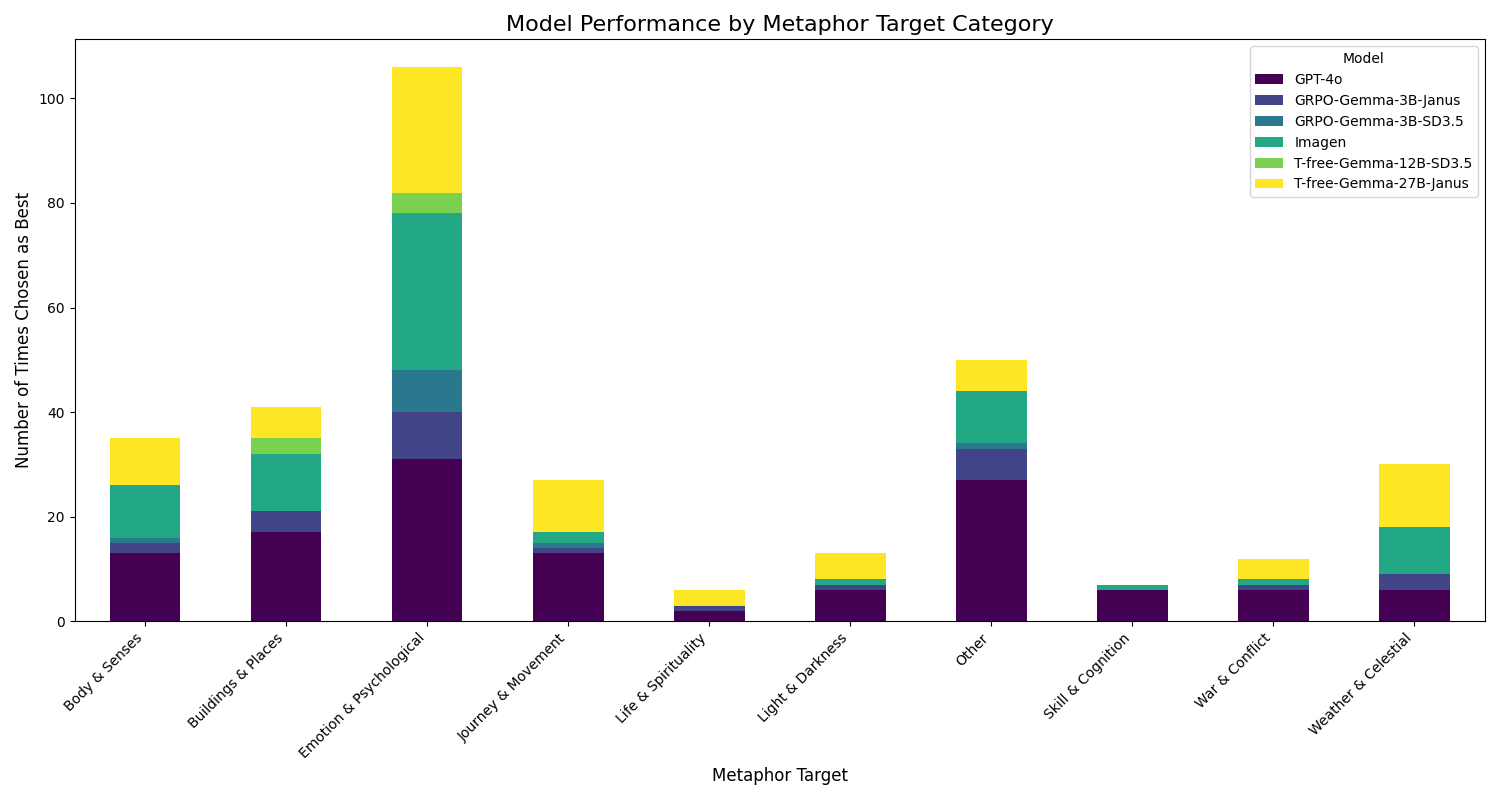}
    \label{fig:target_perf}
  }
  \caption{Model's Performance w.r.t Metaphor Source and Target}
  \label{fig:s_and_t}
\end{figure}

Figures~\ref{fig:source_perf} and \ref{fig:target_perf} decompose the automated evaluation into Source-presence (S-p) and Target-presence (T-p), indicating how clearly each model depicts the concrete source (S) and how effectively it uses S to symbolize the abstract target (T); higher is better. These scores come from the VLM analysis prompts, where S-p measures visibility of S and T-p measures symbolic evocation of T via composition, color, atmosphere, etc.

\begin{figure}[!ht]
  \centering
  \subfigure[Model Performance Compared with Metaphor Length]{
    \includegraphics[width=0.45\textwidth, height=5cm]{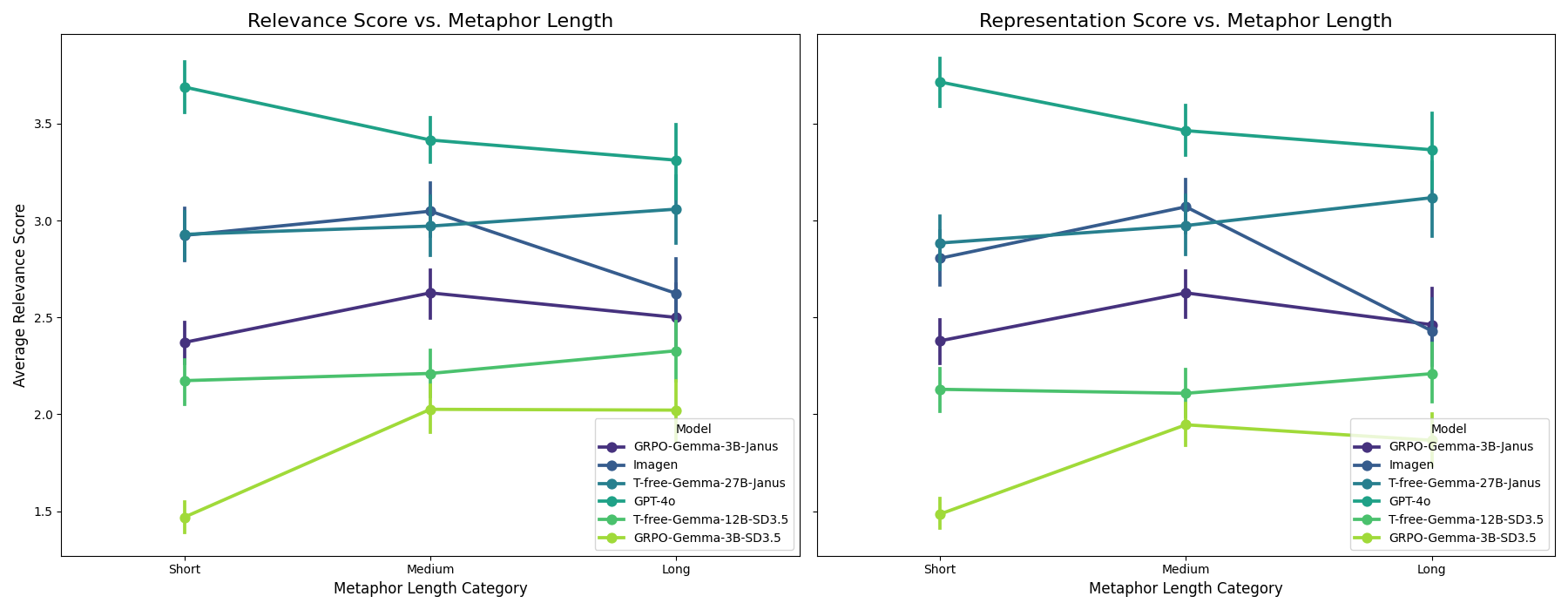}
    \label{fig:model_vs_length}
  }
  \subfigure[Model Win Rates]{
    \includegraphics[width=0.45\textwidth]{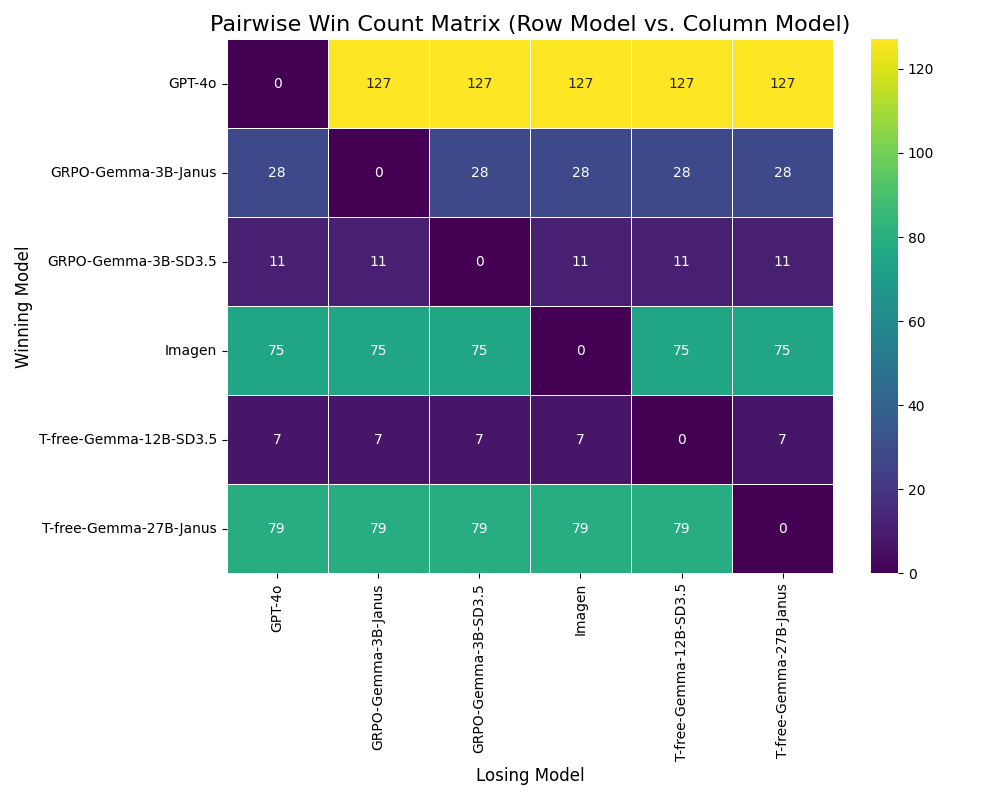}
    \label{fig:win_rate}
  }
  \caption{Performance by Length and Model Win Rates}
  \label{fig:len_and_win_rate}
\end{figure}

Figures~\ref{fig:model_vs_length} and \ref{fig:win_rate} summarize results along two axes: (a) performance stratified by metaphor length (short vs. long), and (b) best-image vote counts from the $15$-annotator user study on $50$ test metaphors, reflecting how often each model’s image was chosen as best overall. The length plot probes robustness to prompt complexity, while the win-rate plot reports aggregate human preference across models.

\section{Prompt Templates}  \label{sec:appendix_templates}

The templates below are used with the VLM model \qwen-32B to evaluate the images generated by zero-shot models \gpt and \imagen. Since \imagen does not have language generation capability, the images are evaluated using `VLM prompt w/o STM'~\ref{subsec:appendix_wo_stm}, whereas \gpt-generated images are evaluated using `VLM prompt w/ STM'~\ref{subsec:appendix_w_stm}.

\subsection{VLM Prompt With STM}    \label{subsec:appendix_w_stm}
\begin{lstlisting}[numbers=none]
VLM_PROMPT_WITH_STM = """
You are an expert art critic. Your task is to analyze an image generated to visualize a metaphor.
The original metaphor is: '{metaphor}'.
The intended breakdown was:
- Source (S): '{s}' (the concrete element)
- Target (T): '{t}' (the abstract concept)
- Meaning (M): '{m}' (the intended connection)

Critically analyze the provided image based on this context. Be strict.
1.  **Perceived Source (S')**: What is the primary visual element in the image that represents '{s}'?
2.  **Perceived Target (T')**: What visual elements, if any, symbolize or evoke the abstract concept of '{t}'? If the image ONLY shows '{s}' without any clear visual link to '{t}', state that the target is not represented.
3.  **Perceived Meaning (M')**: What is the overall meaning the image conveys?
4.  **S Presence Score**: How clearly is '{s}' depicted? (0.0 for not present, 1.0 for clearly present).
5.  **T Presence Score**: How effectively does the image use the visuals of S to symbolize or evoke T? A score of 0.1 should be given if the image only shows S without any metaphorical connection to T. A high score requires clear symbolic elements (e.g., color, composition, atmosphere) that represent T. Be critical.
6.  **Meaning Alignment Score**: Based on your analysis, how well does the Perceived Meaning (M') align with the original intended Meaning (M)? Provide a score from 0.0 (no alignment) to 1.0 (perfect alignment).

Respond with a JSON object with keys 's_prime', 't_prime', 'm_prime', 's_presence_score', 't_presence_score', and 'meaning_alignment_score'.
"""
\end{lstlisting}

\subsection{VLM Prompt Without STM}    \label{subsec:appendix_wo_stm}
\begin{lstlisting}[numbers=none]
VLM_PROMPT_WO_STM = """
You are an expert art critic. Your task is to analyze an image generated to visualize a metaphor.
The original metaphor is: '{metaphor}'.

Critically analyze how well the provided image visually represents this metaphor.
1.  **Visual Description**: Briefly describe the main elements and style of the image.
2.  **Metaphorical Alignment**: How well does the image capture the essence and meaning of the metaphor?
3.  **Alignment Score**: Provide a single score from 0.0 (no connection) to 1.0 (perfectly represents the metaphor) for how well the image visualizes the metaphor.

Respond with a JSON object with keys 'visual_description', 'metaphorical_alignment', and 'alignment_score'.
"""
\end{lstlisting}
\vspace{0.5cm}

The template~\ref{subsec:appendix_decomp} analyses the metaphor decomposition provided by an LLM and provides a decomposition score. The template~\ref{subsec:appendix_vlm}, on the other hand, is used to evaluate the images generated from both the Training-free (Section~\ref{subsec:in-context}) and GRPO (Section~\ref{subsec:grpo}) pipelines.

\subsection{Metaphor Decomposition Analysis}    \label{subsec:appendix_decomp}
\begin{lstlisting}[numbers=none]
f"""
You are an expert in linguistics and semantics. Evaluate the following decomposition of a metaphor.
Original Metaphor: "{metaphor}"

Decomposition:
- Source (S): "{s}"
- Target (T): "{t}"
- Meaning (M): "{m}"

Critique this decomposition. Does 'S' correctly identify the concrete concept? Does 'T' correctly identify the abstract concept? Does 'M' accurately capture the intended connection?

Provide a single score from 0.0 (completely wrong) to 1.0 (perfectly accurate) representing the quality of this decomposition.
Respond using XML-style tags. Put the score inside <decomposition_score> tags and the explanation inside <explanation> tags.

Example:
<decomposition_score>0.8</decomposition_score>
<explanation>The decomposition is mostly correct, but the meaning could be more precise.</explanation>
"""
\end{lstlisting}

\subsection{VLM Analysis Prompt for our pipeline}   \label{subsec:appendix_vlm}
\begin{lstlisting}[numbers=none]
vlm_analysis_prompt = (
    f"You are an expert art critic. Your task is to analyze an image generated to visualize a metaphor. "
    f"The original metaphor is: '{metaphor_full_prompt}'.\n"
    f"The intended breakdown was:\n"
    f"- Source (S): '{s}' (the concrete element)\n"
    f"- Target (T): '{t}' (the abstract concept)\n"
    f"- Meaning (M): '{m}' (the intended connection)\n\n"
    f"Critically analyze the provided image based on this context. Be strict.\n"
    f"1. Perceived Source (S'): What is the primary visual element in the image that represents '{s}'?\n"
    f"2. Perceived Target (T'): What visual elements, if any, symbolize or evoke the abstract concept of '{t}'? If the image ONLY shows '{s}' without any clear visual link to '{t}', state that the target is not represented.\n"
    f"3. Perceived Meaning (M'): What is the overall meaning the image conveys?\n"
    f"4. S Presence Score: How clearly is '{s}' depicted? (0.0 for not present, 1.0 for clearly present).\n"
    f"5. T Presence Score: How effectively does the image use the visuals of S to symbolize or evoke T? A score of 0.1 should be given if the image only shows S without any metaphorical connection to T. A high score requires clear symbolic elements (e.g., color, composition, atmosphere) that represent T. Be critical.\n"
    f"6. Meaning Alignment Score: Based on your analysis, how well does the Perceived Meaning (M') align with the original intended Meaning (M)? Provide a score from 0.0 (no alignment) to 1.0 (perfect alignment).\n\n"
    f"Respond using XML-style tags. Put each answer inside its own tag. Example:\n"
    f"<s_prime>A large, ancient tree.</s_prime>\n"
    f"<t_prime>The concept of 'wisdom' is evoked by the tree's gnarled branches and deep roots.</t_prime>\n"
    f"<m_prime>The image suggests that wisdom is something that grows over a long time and is deeply rooted.</m_prime>\n"
    f"<s_presence_score>0.9</s_presence_score>\n"
    f"<t_presence_score>0.7</t_presence_score>\n"
    f"<meaning_alignment_score>0.8</meaning_alignment_score>"
)
\end{lstlisting}
\vspace{0.5cm}

The template~\ref{subsec:appendix_llm} is used by the LLM to generate a visual prompt whereas the template~\ref{subsec:appendix_refine} is used by the LLM to refine its visual prompt based on the VLM feedback.

\subsection{LLM Prompt for Visual Prompt Generation}    \label{subsec:appendix_llm}
\begin{lstlisting}[numbers=none]
f"""Analyze the following metaphor: "{metaphor}"

Your task is to:
1. Identify the Source (S), Target (T), and Meaning (M) of the metaphor
2. Generate a detailed visual prompt for an image generation model

Instructions:
- S should be the concrete concept used to explain T
- T should be the abstract concept being explained  
- M should be the intended connection or interpretation
- The visual prompt must create a scene that visually represents how T is like S, embodying M
- The visual prompt should be rich in visual details, atmosphere, and composition
- The visual prompt MUST be 77 tokens or less

Please format your response using the following tags:
<reasoning>Your analysis of the metaphor</reasoning>
<source>The concrete source concept</source>
<target>The abstract target concept</target>
<intended_meaning>The metaphorical connection</intended_meaning>
<visual_prompt>A detailed visual description for image generation (<77 tokens)</visual_prompt>"""

    # An example for one-shot learning
    example_user = 'Analyze the following metaphor: "Ideas are diamonds."'
    example_assistant = """<reasoning>The metaphor "Ideas are diamonds" equates ideas with diamonds, suggesting that ideas, like diamonds, are rare, valuable, and formed under pressure. They are initially rough but can be polished to become brilliant and precious.</reasoning>
    <source>Diamonds</source>
    <target>Ideas</target>
    <intended_meaning>Ideas are valuable, rare, and can be refined to brilliance.</intended_meaning>
    <visual_prompt>A brilliant, multifaceted diamond glowing on a dark, velvet cushion. Inside the diamond, intricate neural networks and glowing synapses pulse with light, representing the birth of a powerful idea. The background is dark and abstract, focusing all attention on the diamond's inner light. Cinematic, dramatic lighting.</visual_prompt>"""
\end{lstlisting}

\subsection{Refine LLM Visual Prompt based on VLM Feedback}    \label{subsec:appendix_refine}
\begin{lstlisting}[numbers=none]
decomposition_feedback = f"""
Decomposition Quality Score: {decomposition_quality:.4f}
- This score reflects how well the original S, T, M breakdown captures the metaphor's meaning.
- A low score suggests the decomposition might be inaccurate or incomplete.
"""

    feedback_prompt = f"""\
Original Metaphor: "{original_metaphor}"
  - Source (S): {s}
  - Target (T): {t}
  - Meaning (M): {m}

{decomposition_feedback}

Current Image Generation Prompt: "{current_prompt}"

Evaluation Feedback:
Overall Reward: {reward:.4f}
Individual Scores:
{scores_summary}

Perceived by VLM from the last generated image:
  - Perceived Source (S'): {s_prime if s_prime else 'Not identified'}
  - Perceived Target (T'): {t_prime if t_prime else 'Not identified'}
  - Perceived Meaning (M'): {m_prime if m_prime else 'Not interpreted'}

Task: Based on this feedback, suggest a revised image generation prompt. 
The new prompt should aim to:
1. Better represent the original Source (S), Target (T), and Meaning (M) in the image.
2. Address any weaknesses indicated by the scores (e.g., if S' is different from S, or if M' misaligns with M).
3. If the decomposition quality is low, focus on the most reliable aspects of the S, T, M breakdown.

Provide *only* the new, revised image generation prompt as a single string. Do not include any other explanatory text or labels, and the prompt MUST be 77 tokens or less.
"""
\end{lstlisting}

\section{User Study Instructions}   \label{sec:appendix_user}

We evaluated human preference and metaphor understanding on a subset of test metaphors. For each metaphor, annotators viewed multiple anonymized images generated by different systems. They completed two $5$-point Likert ratings: Metaphor Alignment (how well the image conveys the intended figurative meaning) and Source/Target Representation (how clearly the concrete Source is depicted and how well it evokes the abstract Target), and then selected a single Best Image.

\begin{figure}[!ht]
  \centering
  \subfigure[User study instructions]{
    \includegraphics[width=0.45\textwidth]{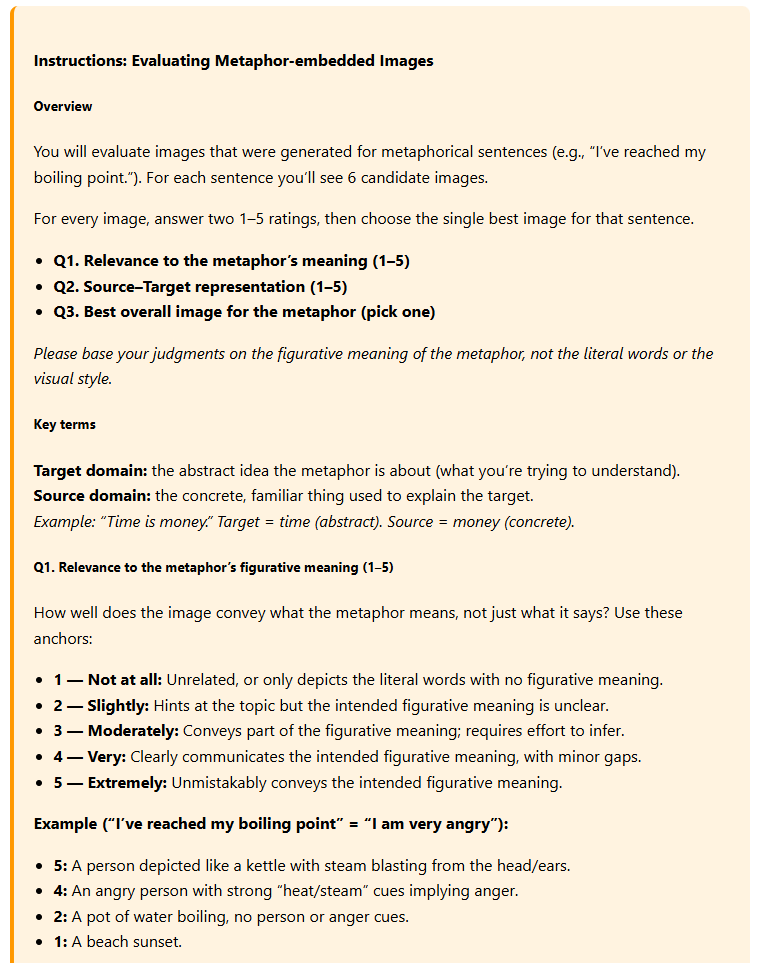}
    \label{fig:page_1}
  }
  \subfigure[User study instructions continued]{
    \includegraphics[width=0.45\textwidth]{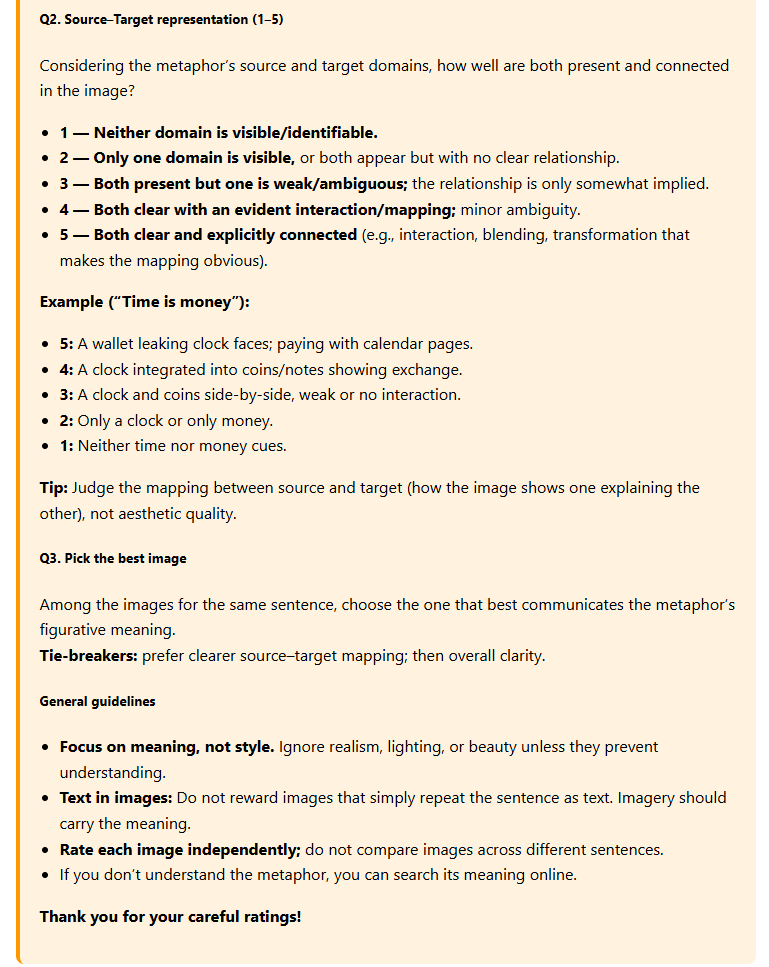}
    \label{fig:page_2}
  }
  \caption{User study instructions provided to the participants.}
  \label{fig:user_instr}
\end{figure}


\newpage
\section*{NeurIPS Paper Checklist}

\begin{enumerate}

\item {\bf Claims}
    \item[] Question: Do the main claims made in the abstract and introduction accurately reflect the paper's contributions and scope?
    \item[] Answer: \answerYes{} 
    \item[] Justification: Section~\ref{sec:experiments} discusses related experiments and results claimed in the abstract and introduction in full detail. The conclusion supports the paper's contributions and scope.

\item {\bf Limitations}
    \item[] Question: Does the paper discuss the limitations of the work performed by the authors?
    \item[] Answer: \answerYes{} 
    \item[] Justification: Section~\ref{subsec:discussion} and the \textbf{Limitations} paragraph discuss them in detail.

\item {\bf Theory assumptions and proofs}
    \item[] Question: For each theoretical result, does the paper provide the full set of assumptions and a complete (and correct) proof?
    \item[] Answer: \answerNA{} 
    \item[] Justification: The paper does not include theoretical results.

    \item {\bf Experimental result reproducibility}
    \item[] Question: Does the paper fully disclose all the information needed to reproduce the main experimental results of the paper to the extent that it affects the main claims and/or conclusions of the paper (regardless of whether the code and data are provided or not)?
    \item[] Answer: \answerYes{} 
    \item[] Justification: Appendix \ref{sec:exp-setup-app} and \ref{sec:appendix_templates} include detailed experimental setup, prompt templates and hyperparameter settings used in our work.

\item {\bf Open access to data and code}
    \item[] Question: Does the paper provide open access to the data and code, with sufficient instructions to faithfully reproduce the main experimental results, as described in supplemental material?
    \item[] Answer: \answerNo{} 
    \item[] Justification: The code will be published upon the paper's acceptance.

\item {\bf Experimental setting/details}
    \item[] Question: Does the paper specify all the training and test details (e.g., data splits, hyperparameters, how they were chosen, type of optimizer, etc.) necessary to understand the results?
    \item[] Answer: \answerYes{} 
    \item[] Justification: Appendix \ref{sec:exp-setup-app} and \ref{sec:appendix_templates} include detailed experimental setup, prompt templates and hyperparameter settings used in our work.

\item {\bf Experiment statistical significance}
    \item[] Question: Does the paper report error bars suitably and correctly defined or other appropriate information about the statistical significance of the experiments?
    \item[] Answer: \answerNo{} 
    \item[] Justification: The statistical significance tests were not carried out since they are not relevant here.

\item {\bf Experiments compute resources}
    \item[] Question: For each experiment, does the paper provide sufficient information on the computer resources (type of compute workers, memory, time of execution) needed to reproduce the experiments?
    \item[] Answer: \answerYes{} 
    \item[] Justification: The paper includes the computational resources required to run the experiments in Appendix \ref{sec:exp-setup-app}.
    
\item {\bf Code of ethics}
    \item[] Question: Does the research conducted in the paper conform, in every respect, with the NeurIPS Code of Ethics \url{https://neurips.cc/public/EthicsGuidelines}?
    \item[] Answer: \answerYes{} 
    \item[] Justification: The paper conforms with the Code of Ethics.

\item {\bf Broader impacts}
    \item[] Question: Does the paper discuss both potential positive societal impacts and negative societal impacts of the work performed?
    \item[] Answer: \answerNA{} 
    \item[] Justification: The paper mainly focuses on the creative abilities of generative AI models and is not related to the societal impact.
    
\item {\bf Safeguards}
    \item[] Question: Does the paper describe safeguards that have been put in place for responsible release of data or models that have a high risk for misuse (e.g., pretrained language models, image generators, or scraped datasets)?
    \item[] Answer: \answerNA{} 
    \item[] Justification: The paper does not release models or datasets.

\item {\bf Licenses for existing assets}
    \item[] Question: Are the creators or original owners of assets (e.g., code, data, models), used in the paper, properly credited and are the license and terms of use explicitly mentioned and properly respected?
    \item[] Answer: \answerYes{} 
    \item[] Justification: Existing assets used in the paper are credited and adhere to the licenses mentioned.

\item {\bf New assets}
    \item[] Question: Are new assets introduced in the paper well documented and is the documentation provided alongside the assets?
    \item[] Answer: \answerNA{} 
    \item[] Justification: No new assets are introduced.

\item {\bf Crowdsourcing and research with human subjects}
    \item[] Question: For crowdsourcing experiments and research with human subjects, does the paper include the full text of instructions given to participants and screenshots, if applicable, as well as details about compensation (if any)? 
    \item[] Answer: \answerYes{} 
    \item[] Justification: The user study is discussed in Section~\ref{subsec:user-study} and the participant instructions are outlined in Appendix~\ref{sec:appendix_user}.

\item {\bf Institutional review board (IRB) approvals or equivalent for research with human subjects}
    \item[] Question: Does the paper describe potential risks incurred by study participants, whether such risks were disclosed to the subjects, and whether Institutional Review Board (IRB) approvals (or an equivalent approval/review based on the requirements of your country or institution) were obtained?
    \item[] Answer: \answerNA{} 
    \item[] Justification: There are no potential risks; however, ethics approval was taken from the organization before the study.

\item {\bf Declaration of LLM usage}
    \item[] Question: Does the paper describe the usage of LLMs if it is an important, original, or non-standard component of the core methods in this research? Note that if the LLM is used only for writing, editing, or formatting purposes and does not impact the core methodology, scientific rigorousness, or originality of the research, declaration is not required.
    \item[] Answer: \answerYes{} 
    \item[] Justification: The paper is about the creative abilities of generative AI models.

\end{enumerate}

\end{document}